\documentclass{article} 
\usepackage{nips13submit_e,times}
\usepackage{url}
\usepackage{graphicx}
\usepackage{epstopdf}
\usepackage{longtable}
\usepackage{tabularx}
\usepackage{amsfonts}
\usepackage{amsmath}
\usepackage{mathtools}
\usepackage{fancyvrb}
\usepackage{pdflscape}
\usepackage{multirow}
\usepackage{enumitem}
\usepackage{booktabs}

\title{Deep Learning for Answer Sentence Selection}

\author{
Lei Yu$^1$\ \ \ \ \ Karl Moritz Hermann$^2$\ \ \ \ \ Phil Blunsom$^{12}$\ \ \ \ \ Stephen Pulman$^1$\\
$^1$Department of Computer Science, University of Oxford\\
$^2$Google DeepMind \\
\texttt{\{lei.yu,\,phil.blunsom,\,stephen.pulman\}@cs.ox.ac.uk}\\
\texttt{kmh@google.com}
}

\nipsfinalcopy 

\begin{document}

\maketitle

\begin{abstract}
Answer sentence selection is the task of identifying sentences that contain the
answer to a given question. This is an important problem in its own right as
well as in the larger context of open domain
question answering. We propose a novel approach to solving this task via
means of distributed representations, and learn to match questions with answers by
considering their semantic encoding.  This contrasts prior work on this task,
which typically relies on classifiers with large numbers of hand-crafted
syntactic and semantic features and various external resources. Our approach 
does not require any feature engineering nor does it involve specialist
linguistic data, making this model easily applicable to a wide range of domains
and languages.  Experimental results on a standard benchmark dataset from TREC
demonstrate that---despite its simplicity---our model matches state of the art
performance on the answer sentence selection task.
\end{abstract}

\section{Introduction}

Question answering can broadly be divided into two categories. One approach
focuses on semantic parsing, where answers are retrieved by turning a question
into a database query and subsequently applying that query to an existing
knowledge base. The other category is open domain question answering, which is
more closely related to the field of information retrieval.

Open domain question answering requires a number of intermediate steps. For
instance, to answer a question such as {\it ``Who wrote the book Harry Potter?''}, a
system would first identify the question type and retrieve relevant documents.
Subsequently, within the retrieved documents, a sentence containing the answer
is selected, and finally the answer ({\it J.K. Rowling}) itself is extracted
from the relevant sentence.
In this paper, we focus on answer
sentence selection, the task that selects the correct sentences answering a
factual question from a set of candidate sentences. Beyond its role in open
domain question answering, answer sentence selection is also a stand-alone task
with applications in knowledge base construction and information extraction.
The correct sentence may not
answer the question directly and perhaps also contain extraneous information,
for example:

\begin{enumerate}
\item[]\textbf{Q:} When did Amtrak begin operations?
\item[]\textbf{A:} Amtrak has not turned a profit since it was founded in 1971.
\end{enumerate}

The relevance of an answer sentence to a question is typically determined by
measuring the semantic similarity between question and answer.
Prior work in this field mainly attempted this via syntactic matching of parse
trees. This can be achieved with generative models that syntactically transform 
answers to questions \cite{wang2007jeopardy,wang2010probabilistic}. Another 
option is discriminative models over features produced from minimal edit sequences 
between dependency parse trees \cite{DBLP:conf/naacl/HeilmanS10a,yao2013answer}.
Beyond syntactic information, some prior work has also included semantic
features from resources such as WordNet, with the previous state-of-the-art
model for this task relying on a variety of such lexical semantic resources
 \cite{yih2013question}.
While empirically the inclusion of large amounts of semantic information has
been shown to improve performance, such approaches rely on a significant amount
of feature engineering and require expensive semantic resources
which may be difficult to obtain, particularly for resource-low languages.
A second limitation of such feature-based semantic models is the difficulty of
adapting to new domains, requiring separate feature
extraction and resource development or identification steps for every domain.

At the same time, neural network-based distributional sentence models have
achieved successes in many natural language processing tasks such as sentiment
analysis \cite{DBLP:conf/acl/HermannB13,DBLP:conf/acl/KalchbrennerGB14},
paraphrase detection \cite{DBLP:conf/emnlp/SocherHMN12} and document
classification \cite{Hermann:2014:ACLphil,Hermann:2014:ICLR}.

As a consequence of this success, it appears natural to attempt to solve question
answering using similar techniques. There has been some work on this in the
context of semantic parsing for knowledge base question answering
\cite[\textit{inter alia}]{DBLP:conf/acl/YihHM14,DBLP:journals/corr/BordesWU14},
where the focus is on learning semantic representations for knowledge base
tuples. Another line of work---closely related to the model presented here---is
the application of recursive neural networks to factoid question answering over
paragraphs \cite{Iyyer:Boyd-Graber:Claudino:Socher:Daume-2014}.  A key
difference to our approach is that this model, given a question, selects answers
from a relatively small fixed set of candidates encountered during training. On the
other hand, the task of answer sentence selection that we address here, requires
picking an answer from among a set of candidate sentences not encountered during
training. In addition, each question has different numbers of candidate sentences.

In this paper, we show that a neural network-based sentence model can be applied
to the task of answer sentence selection. We construct two distributional
sentence models; first a bag-of-words model, and second, a bigram model
based on a convolutional neural network. Assuming a set of pre-trained semantic
word embeddings, we train a supervised model to learn a semantic
matching between question and answer pairs. The underlying distributional models
provide the semantic information necessary for this matching function. We also
present an enhanced version of this model, which combines the signal of the
distributed matching algorithm with two simple word matching features. Note that
these features still do not require any external linguistic annotation.

Compared with previous work on answer sentence selection, our approach
is applicable to any language, and  does not require feature-engineering 
and hand-coded resources beyond some large corpus on which to train
our initial word embeddings.
We conduct experiments on an answer selection dataset created from
TREC QA track. The results show that our models are very effective on
this task --- matching the state-of-the-art results.

\section{Related work}

There are three threads of related work relevant to our approach. We first
review composition methods for distributional semantics and then
discuss the existing work on answer sentence selection. Finally, we introduce
existing work on applying neural networks to question answering.

{\bf Compositional distributional semantics.} Compared with words in string
form or logical form, distributed representations of words can capture latent 
semantic information and thereby exploit similarities between words. This 
enables distributed representations to overcome sparsity problems encountered 
in atomic representations and further provides information about how words 
are semantically related to each other. These properties have caused distributed 
representations to become increasingly popular in natural language processing.
They have been proved to be successful in applications such as relation extraction \cite{Grefenstette:1994:EAT:527911} and word sense disambiguation 
\cite{DBLP:conf/acl/McCarthyKWC04}. Vector representations for words can be 
obtained in a number of ways, with many approaches exploiting distributional 
information from large corpora, for instance by counting the frequencies of 
contextual words around a given token. An alternative method for learning 
distributed representations comes in the form of neural language models, where 
the link to distributional data is somewhat more obscure
\cite{DBLP:conf/icml/CollobertW08,DBLP:conf/nips/BengioDV00}. A nice side-effect
of the way in which neural language models learn word embeddings is that these
vectors implicitly contain both syntactic and semantic information.

For a number of tasks, vector representations of single words are not
sufficient. Instead, distributed representations of phrases and
sentences are required in order to provide a deeper language understanding that
allows addressing more complex tasks in NLP such as translation, sentiment
analysis or information extraction.
As sparsity prevents us from directly learning distributional representations at
the phrase level, various models of vector composition have been proposed that
circumvent this problem by learning higher level representations based on
low-level (e.g. word-level) representations.
Popular ideas for this include exploiting category theory
\cite{ClarkCoeckeSadrzadeh2008}, using parse trees in conjunction with recursive
autoencoders \cite{DBLP:conf/emnlp/SocherHMN12,DBLP:conf/acl/HermannB13}, and
convolutional neural networks
\cite{DBLP:conf/icml/CollobertW08,DBLP:conf/acl/KalchbrennerGB14}. More
recently, sentence models constructed from parallel corpora
\cite{Hermann:2014:ACLphil} point to a new trend in compositional
distributional semantics, where word- and sentence-level representations are
learned given a joint semantic objective function.

{\bf Answer sentence selection.} Answer sentence selection denotes the task
of selecting a sentence that contains the
information required to answer a given question from a set of candidates
obtained via some information extraction system.

Clearly, answer sentence selection requires both semantic and syntactic
information in order to establish both what information the question seeks to
answer, as well as whether a given candidate contains the required information,
with current state-of-the-art approaches mostly focusing on syntactic matching
between questions and answers.
Following the idea that questions can be generated from correct answers by
loose syntactic transformations, Wang et al. \cite{wang2007jeopardy} built a
generative model to match the dependency trees of question answer pairs based on
the soft alignment of a quasi-synchronous grammar \cite{smith2006quasi}.
Wang and Manning \cite{wang2010probabilistic} proposed another probabilistic
model based on Conditional Random Fields, which models alignment as a set of
tree-edit operations of dependency trees. Heilman and Smith
\cite{DBLP:conf/naacl/HeilmanS10a} used a tree kernel as a heuristic to search
for the minimal edit sequences between parse trees. Features extracted from
these sequences are then fed into a logistic regression classifier to select the
best candidate.
More recently, Yao et al. \cite{yao2013answer} extended Heilman and Smith's
approach with the difference that they used dynamic programming to find the
optimal tree edit sequences. In addition, they added semantic features obtained from WordNet.
 Although some of these approaches use
WordNet relations (e.g. synonym, antonym, hypernym) as explicit features, the
focus of all of this work is primarily on syntactic information
\cite{yih2013question}.

Unlike previous work, Yih et al. \cite{yih2013question} applied rich lexical
semantics to their state-of-the-art QA matching models. These models match
the semantic relations of aligned words in QA pairs by using a combination of
lexical semantic resources such as WordNet with distributional representations
for capturing semantic similarity. This approach results in a series of features
for sentence pairs, which are then fed into a conventional classifier. A
variation of this idea can also be found in Severyn and Moschitti
\cite{severynautomatic}, who used an SVM with tree kernels to automatically learn
features from shallow parse trees rather than relying on external resources,
sacrificing semantic information for model simplicity. This paper combines these
two approaches by proposing a semantically rich
model without the need for feature engineering or extensive human-annotated 
external resources.

{\bf Applying neural networks to question answering.} Only very recently have
researchers started to apply deep learning to question
answering. Relevant work includes Yih et al.~\cite{DBLP:conf/acl/YihHM14}, who
constructed models for single-relation question answering with a knowledge base of triples.
In the same direction, Bordes et
al.~\cite{DBLP:journals/corr/BordesWU14,Bordes:2014:EMNLP} used a type of siamese
network for learning to project question and answer pairs into a joint space.
Finally, Iyyer et al.~\cite{Iyyer:Boyd-Graber:Claudino:Socher:Daume-2014} worked
on the quiz bowl task, a question answering task that requires identifying an
entity as described by a series of sentences. They modelled semantic composition
with a recursive neural network.
Both of these tasks differ from the work presented here in that answer selection
can require mapping questions containing multiple relations to answer sentences
also containing several concepts and relations.

\section{Model description}\label{model}

Answer sentence selection can be viewed as a binary classification problem.
Assume a set of questions $Q$, where each question $\mathbf{q}_i \in Q$ is
associated with a list of answer sentences $\{\mathbf{a}_{i1}, \mathbf{a}_{i2},
\cdots, \mathbf{a}_{im}\}$, together with their judgements $\{y_{i1}, y_{i2},
\cdots, y_{im}\}$, where $y_{ij} = 1$ if the answer is correct and $y_{ij} =
0$ otherwise. While this could be approached as a multi-labelling task, we
simply treat each data point as a triple $(\mathbf{q}_i, \mathbf{a}_{ij},
  y_{ij})$. Thus, our task is to learn a classifier over these triples so
that it can predict the judgements of any additional QA pairs.

Our solution to this problem assumes that correct answers have high
semantic similarity to questions. Unlike previous work, which measured the
similarity mainly using syntactic information and hand-crafted semantic
resources, we model questions and answers as vectors, and evaluate the
relatedness of each QA pair in a shared vector space. Formally, following Bordes
et al. \cite{DBLP:journals/corr/BordesWU14}, given the vector representations of
a question $\mathbf{q}$ and an answer $\mathbf{a}$ (both in $\mathbb{R}^{d}$), 
the probability of the answer being correct is
\begin{equation}
p(y = 1 | \mathbf{q}, \mathbf{a}) = \sigma(\mathbf{q}^T\ \mathbf{M}\ \mathbf{a}
  + b),\\[0.5em]
\end{equation}
where the bias term $b$ and the transformation matrix $\mathbf{M} \in
\mathbb{R}^{d \times d}$ are model parameters.  This formulation can intuitively
be understood as an expression of the generative approach to open domain
question answering: given a candidate answer sentence, we `generate' a question
through the transformation $\mathbf{q}' = \mathbf{M}\ \mathbf{a}$, and
then measure the similarity of the generated question $\mathbf{q}'$ and the
given question $\mathbf{q}$ by their dot product. The sigmoid function
squashes the similarity scores to a probability between $0$ and $1$. The model
is trained by minimising the cross entropy of all labelled data QA pairs:
\begin{equation}
\begin{split}
\mathcal{L} &= -\log \prod_n p(y_n | \mathbf{q}_n, \mathbf{a}_n) + \frac{\lambda}{2}\|\theta\|^2_F\\
& = -\sum_n y_n \log \sigma(\mathbf{q}_n^T\ \mathbf{M}\ \mathbf{a}_n + b) + (1 - y_n)\log (1 - \sigma(\mathbf{q}_n^T\ \mathbf{M}\ \mathbf{a}_n + b)) + \frac{\lambda}{2}\|\theta\|^2_F,
\end{split}
\end{equation}
where $\|\theta\|^2_F$ is the Frobenius norm of $\theta$, and $\theta$ includes
$\{\mathbf{M},b\}$ as well as any parameters introduced in the sentence
composition  model. Next we describe two methods employed in this work for
projecting sentences into vector space representations.

\subsection{Bag-of-words model}

Given word embeddings, the bag-of-words model generates the vector
representation of a sentence by summing over the embeddings of all words in the
sentence---having previously removed stop words from the input. The vector is
then normalised by the length of the sentence.
\begin{equation}
\mathbf{s} = \frac{1}{|\mathbf{s}|}\sum_{i=1}^{|\mathbf{s}|} \mathbf{s}_i.
\end{equation}

\subsection{Bigram model}

Due to its inability to account for word ordering and other structural
information, the simple bag-of-words model proposed above is unable to capture
more complex semantics of a sentence. To address this issue, we also evaluate a
sentence model based on a convolutional neural network (CNN).

The advantage of this composition model is that it is sensitive to word ordering
and is able to capture features of $n$-grams independent of their positions in
the sentences. Further, the convolutional network can learn to correspond to the
internal syntactic structure of sentences, removing reliance on external
resources such as parse trees \cite{DBLP:conf/acl/KalchbrennerGB14}. Finally,
the convolution and pooling layers help us to capture long-range dependencies,
which are common in questions. CNN-based models have been proved to be effective
in applications such as semantic role labelling
\cite{DBLP:conf/icml/CollobertW08}, twitter sentiment prediction
\cite{DBLP:conf/acl/KalchbrennerGB14} and semantic parsing
\cite{DBLP:conf/acl/YihHM14}.
\begin{figure}[h]
\begin{center}
\includegraphics[scale = 0.4]{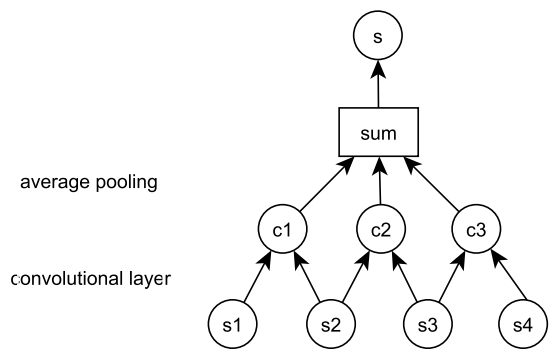}
\end{center}
\caption{The architecture of a one-dimensional convolutional neural network}
\label{ccn}
\end{figure}
Figure \ref{ccn} illustrates the architecture of the CNN-based sentence model in
one dimension. We use bigrams here with one convolutional layer and one pooling
layer. The convolutional vector $\mathbf{t} \in \mathbb{R}^2$, which is shared
by all bigrams, projects every bigram into a feature value $\mathbf{c}_i$,
computed as follows:
\begin{equation}
\mathbf{c}_i = \tanh(\mathbf{t}\cdot\mathbf{s}_{i:i+1} + b).
\label{convolve}
\end{equation}
Since we would like to capture the meaning of the full sentence, we then use
average pooling to combine all bigram features. This produces a full-sentence
representation of the same dimensionality as the initial word embeddings. In
practice, of course, these representations are not a single value, but
$d$-dimensional vectors, and hence $\mathbf{t}$ and each bigram are matrices.
Thus, Equation \ref{convolve} is calculated for each row of $\mathbf{t}$ and the
corresponding row of $\mathbf{s}$. Similarly, average pooling is performed
across each row of the convolved matrix. Formally, our bigram model is
\begin{equation}
\mathbf{s} = \sum_{i=1}^{|\mathbf{s}|-1} \tanh (\mathbf{T}_L\ \mathbf{s}_{i} +
  \mathbf{T}_R\ \mathbf{s}_{i+1} + \mathbf{b}),
\end{equation}
where $\mathbf{s}_i$ is the vector of the $i$-th word in the sentence, and
$\mathbf{s}$ is the vector representation of the sentence. Both vectors are in
$\mathbb{R}^{d}$. $\mathbf{T}_L$ and $\mathbf{T}_R$ are model
parameters in $\mathbb{R}^{d \times d}$ and $\mathbf{b}$ is bias. From a linguistic perspective, these
parameters can be considered as informing the ordering of individual word pairs.

\section{Experiments}

We evaluated both models presented in this paper on a standard answer selection
dataset. We briefly introduce the dataset before describing our experimental
setup. Finally, we report our results and compare them with previous work.

\subsection{TREC Answer Selection Dataset}

The answer sentence selection dataset contains a set of factoid questions, with
a list of answer sentences corresponding to each question. Wang et al.
\cite{wang2007jeopardy} created this dataset from Text REtrieval Conference
(TREC) QA track ($8$-$13$) data, with candidate answers automatically selected
from each question's document pool. Answer candidates were chosen using a
combination of overlapping non-stop word counts and pattern matching.
Subsequently, the correctness of the candidate answers were judged manually for
parts of the dataset. Table~\ref{data-table} summarises the answer selection 
dataset, and describes the train/dev/test split of the data. The TRAIN-ALL
training set---significantly larger than any of the other parts of the data---is
noisy as answers were labelled automatically using pattern matching.

\begin{table}[t]\centering
  \begin{tabular}{@{}lrrrl@{}}
    \toprule
    Data & \# Questions & \# QA Pairs & \% Correct & Judgement
    \\
    \midrule
    TRAIN-ALL  & 1,229 & 53,417 & 12.0 & automatic \\
    TRAIN	   & 94   & 4,718  & 7.4  & manual \\
    DEV		   & 82	  & 1,148  & 19.3 & manual \\
    TEST	   & 100  & 1,517  & 18.7 & manual \\
    \bottomrule
  \end{tabular}
  \caption{Summary of the answer sentence selection dataset. Judgement denotes
    whether correctness was determined automatically or by human annotators.}
  \label{data-table}
\end{table}

The task is to rank the candidate answers based on their relatedness to the
question, and is thus measured in Mean Average Precision (MAP) and Mean
Reciprocal Rank (MRR), which are standard metrics in Information Retrieval and
Question Answering. Whereas MRR measures the rank of any correct answer,  
MAP examines the ranks of all the correct answers. In general, MRR is slightly higher
than MAP on the same list of ranked outputs, except that they are the
same in the case where each question has exactly one correct answer.
The scores are calculated using the official
\texttt{trec\_eval} evaluation scripts.

\subsection{Experimental setup}

We used word embeddings ($d = 50$) that were computed using Collobert and
Weston's neural language model \cite{DBLP:conf/icml/CollobertW08} and provided
by Turian et al. \cite{DBLP:conf/acl/TurianRB10}. Even though our objective
function would allow us to learn word embeddings directly, we fixed those
representations in light of the small size of the QA answer selection dataset.
The other model weights were randomly intitialised using a Gaussian distribution
($\mu = 0, \sigma = 0.01$).
All hyperparameters were optimised via grid
search on the MAP score on the development data. We use the AdaGrad algorithm
\cite{DBLP:journals/jmlr/DuchiHS11} for training.

One weakness of the distributed approach is that---unlike symbolic
approaches---distributed representations are not very well equipped for dealing
with cardinal numbers and proper nouns, especially considering the small dataset.
As these are important artefacts in any
question answering task, we mitigate this issue by using a second feature that
counts co-occurring words in question-answer pairs. We integrate this feature
with our distributional models by training a logistic regression classifier with
three features: word co-occurrence count, word co-occurrence count weighted by
IDF value and QA matching probability as provided by the distributional model.
Notice that here we did not add more
$n$-gram features, since we would like to see how much the distributional models
contributed in this task, rather than $n$-gram overlapping. L-BFGS was used to
train the logistic regression classifier, with L2 regulariser of $0.01$.

\subsection{Results}

\begin{table}[t]\centering

  \begin{tabular}{@{}llrr@{}}
    \toprule
    \multicolumn{2}{@{}l}{Model} & MAP & MRR \\
    \midrule
    \multicolumn{2}{@{}l}{\textbf{TRAIN}} &&  \\
    & unigram 			& 0.5387 & 0.6284 \\
    & bigram	 		& 0.5476 & 0.6437 \\
    & unigram + count 	& 0.6889 & 0.7727 \\
    & bigram + count	& \bf{0.7058} & \bf{0.7800} \\
    \midrule
    \multicolumn{2}{@{}l}{\textbf{TRAIN-ALL}} &&  \\
    & unigram			& 0.5470 & 0.6329 \\
		& bigram			& 0.5693 & 0.6613 \\
		& unigram + count	& 0.6934 & 0.7677 \\
  & bigram + count	& \bf{0.7113} & \bf{0.7846} \\
    \bottomrule
\end{tabular}
\caption{Results of our models (unigram and bigram) trained on both the TRAIN
  and TRAIN-ALL dataset. \textit{count} indicates whether the overlapping word
  count features were also used.}
\label{our-table}
\end{table}

\begin{table}[t]\centering
  \begin{tabular}{@{}llrr@{}}
    \toprule
    \multicolumn{2}{@{}l}{System} & MAP & MRR \\
    \midrule
    \multicolumn{2}{@{}l}{\textbf{Baselines}} &&  \\
    & Random 		& 0.3965 & 0.4929 \\
    & Word Count  & 0.5707 & 0.6266 \\
    & Wgt Word Count & 0.5961 & 0.6515 \\
    \midrule
    \multicolumn{2}{@{}l}{\textbf{Published Models}} &&  \\
    & Wang et al. (2007)  		& 0.6029 	& 0.6852 \\
    & Heilman and Smith (2010) 	& 0.6091   	& 0.6917 \\
    & Wang and Manning (2010)   & 0.5951    & 0.6951 \\
    & Yao et al. (2013) 			& 0.6307	& 0.7477 \\
    & Severyn and Moschitti (2013)& 0.6781  	& 0.7358 \\
    & Yih et al. (2013) -- LR		& 0.6818	& 0.7616 \\
    & Yih et al. (2013) -- BDT		& 0.6940	& \bf{0.7894} \\
    & Yih et al. (2013) -- LCLR	& \bf{0.7092}	& 0.7700 \\
    \midrule
    \multicolumn{2}{@{}l}{\textbf{Our Models}} &&  \\
    & TRAIN bigram + count	& 0.7058 & 0.7800 \\
    & TRAIN-ALL bigram + count	& \bf{0.7113} & \bf{0.7846} \\
    \bottomrule
  \end{tabular}
  \caption{Overview of results on the QA answer selection task. Baseline models
    are taken from Table 2 in Yih et al. \cite{yih2013question}. We also include
    the results of our best models (bigram + count), which match the current
    state of the art.}
  \label{result-table}
\end{table}

Table \ref{our-table} summarises the results of our models.  As can be seen, the
bigram model performs better than the unigram model and the addition of the
IDF-weighted word count features significantly improve performance for both
models by $10 \%$ -- $15\%$.
Wang et al. \cite{wang2007jeopardy} reported that training with the noisy
dataset TRAIN-ALL negatively impacted their models. This does not apply to our
models, where performance of the models increases.

Table \ref{result-table} surveys published results on this task, and places our
best models in the context of the current state-of-the-art results. The table
also includes three baseline models provided in \cite{yih2013question}. The
first model randomly assigns scores to each answer. The second model counts the
number of words co-occurring in each QA pair,
with another version of that baseline weighting these word counts by IDF values.
As can be seen in Table \ref{result-table}, our best models (bigram + count)
outperform all baselines and prior work on MAP and are very close to the best model
proposed by Yih et al. on MRR. Considering the lack of
complexity of our models compared to those of Yih et al., these results are very
promising and indicate the soundness of our approach to the QA answer selection
task. 

\section{Discussion}

As already stated in the background section of this paper, most prior work
focuses on syntactic analysis, with semantic aspects mainly being incorporated
through a number of manually engineered features and external resources such as
WordNet. Interestingly, however, the best performing published model is also the
only piece of prior work that is primarily focused on semantics
\cite{yih2013question}. In their model, Yih et al. match aligned words between
questions and answers and extract features from word pairs. The word-level
features are then aggregated to represent sentences, which are used for
classification. They combine a group of word matching features with semantic
features obtained from a wide range of linguistic resources including WordNet,
polarity-inducing latent semantic analysis (PILSA) model \cite{yih2012polarity}
and different vector space models.

When considering the heavy reliance of resources of those models in comparison 
to the simplicity of our approach, the relative performance of our models is 
highly encouraging. As we only use two non-distributional features---question-answer
pair word matching and word matching weighted by IDF values---it is plausible to regard
the distributional aspect of our models as a replacement for the numerous
lexical semantic resources and features utilised in Yih et al.'s work.
In the context of these results, it is also worth noting that---unlike the model 
of Yih et al.---our models can directly be applied across languages, as we are not 
relying on any external resources beyond some large corpus on which to train our 
initial word embeddings.

Earlier in this paper we argued that methods based purely on vector
representations may not be sufficient for solving complex problems such as
paraphrase detection and question answering because of their weakness in dealing with
certain aspects of language such as numbers and---to a lesser extent---proper
nouns. For example, the mismatching of numbers are crucial for rejecting a pair of
`paraphrases' or an answer. Surface-form matching is particularly important in
our experiment because we did not learn word embeddings from the answer
selection dataset and the given word dictionaries may not cover all the words in
the dataset\footnote{Approximately $5 \%$ of words in the answer selection
dataset are not covered in Collobert and Weston's embeddings.}. Most of the
non-covered words are proper nouns, which are then assigned the \texttt{UNKNOWN}
token. While these are likely to be crucial for judging the relevance of an
answer candidate, they cannot be incorporated into the distributional aspect of
our model.

However, it is also important to then establish the opposite fact, namely that
distributional semantics improve over purely word counting model.  When
reviewing the baseline results (Table \ref{result-table}) relative to the
performance of our models, it is also evident, that adding distributional
semantics as a feature improves over models based purely on co-occurrence counts
and word matching. In fact, this addition boosts both MAP and MRR scores by
approximately $10\%$.  We analysed this effect by considering sentences where
our combined model (bigram + count) performs better than the counting baseline.
Here are two examples, where the model of co-occurrence word count failed to identify the
correct answer, but the combined model prevailed:

\begin{enumerate}
  \item
    \begin{itemize}[leftmargin=5mm]
      \item[\textbf{Q:}] When  did James Dean {\it die}?
      \item[\textbf{A1:}] In $\langle$num$\rangle$, actor James Dean was {\it
          killed} in a two-car collision near Cholame, Calif. ({\bf correct})
      \item[\textbf{A2:}] In $\langle$num$\rangle$, the studio asked him to
        become a technical adviser on Elia Kazan's ``East of Eden,'' starring
        James Dean. ({\bf incorrect})
    \end{itemize}
  \item
    \begin{itemize}[leftmargin=5mm]
      \item[{\bf Q:}] How many members are there in the {\it singing group} the
        Wiggles?
      \item[{\bf A1:}] The Wiggles are four effervescent {\it performers} from 
        the Sydney area: Anthony Field, Murry Cook, Jeff Fatt and Greg Page. ({\bf correct})
      \item[{\bf A2:}] Let's now give a welcome to the Wiggles, a goofy new
        import from Australia. ({\bf incorrect})
    \end{itemize}
\end{enumerate}

In both cases, the baseline model cannot tell the difference between the two
candidate answers since they have the same number of matched words to the
question. However, for the first example the combined model assigns higher score
to the first answer since the word {\it die} is semantically close to {\it
killed}. Similarly, for the second example, the word {\it performers} in the
first sentence is related to {\it singing group} in the question, and hence the
first one gets higher score.

Although the use of distributional semantics model could bring benefits to pure
syntactic matching, there are still problems that the current method cannot
tackle. For example,

\begin{enumerate}
    \setcounter{enumi}{2}
  \item
    \begin{itemize}[leftmargin=5mm]
      \item[{\bf Q:}]  What is the name of Durst's group?
      \item[{\bf A:}] Limp Bizkit lead singer Fred Durst did a lot before he hit the big time.
    \end{itemize}
\end{enumerate}

Here, the judgement requires a good understanding of the sentence and some
background knowledge about the relation between {\it lead singer} and {\it
group}.

\section{Conclusion}

This paper demonstrated the effectiveness of applying distributional sentence
models to answer sentence selection. We projected questions and answers into 
vectors and learned a semantic matching function between QA pairs. 
Subsequently we combined this function with a simple, weighted QA co-occurrence counter.

We demonstrated that this approach with a bag-of-words sentence model
significantly improves performance over the original count-based baseline model.
By using a more complex sentence model based on a convolutional neural
network over bigrams, we improved performance further still, and attained
the state of the art on the answer selection task.
Compared with previous work based on feature engineering and external
hand-coded semantic resources, our approach is much simpler and more flexible.

In the future, we would like to investigate more complex sentence models for
this task, for example a convolutional network based sentence model with higher order 
$n$-grams and multiple feature maps, as well as recursive neural network-based models.  
Moreover, since answer sentence selection is similar to textual
entailment and paraphrase detection, we are hoping to extend this line of work
to these tasks.

\section*{Acknowledgment}
We thank Wen-tau Yih and Xuchen Yao for sharing the answer sentence selection
dataset.  
This work was supported by a Xerox Foundation
Award and EPSRC grant number EP/K036580/1.

\bibliographystyle{plain}
\bibliography{ref}

\end{document}